Tech Science Press

# Deep Learning-Based 3D Instance and Semantic Segmentation: A Review

**Siddiqui Muhammad Yasir[1] and Hyunsik Ahn[2,*]**

[1]Department of Robot System Engineering, Tongmyong University, Busan, 48520, Korea
[2]School of Artificial Intelligence, Tongmyong University, Busan, 48520, Korea
*Corresponding Author: Hyunsik Ahn. Email: hsahn@tu.ac.kr

**Abstract:** The process of segmenting point cloud data into several homogeneous areas with points in the same region having the same attributes is known as 3D segmentation. Segmentation is challenging with point cloud data due to substantial redundancy, fluctuating sample density and lack of apparent organization. The research area has a wide range of robotics applications, including intelligent vehicles, autonomous mapping and navigation. A number of researchers have introduced various methodologies and algorithms. Deep learning has been successfully used to a spectrum of 2D vision domains as a prevailing A.I. methods. However, due to the specific problems of processing point clouds with deep neural networks, deep learning on point clouds is still in its initial stages. This study examines many strategies that have been presented to 3D instance and semantic segmentation and gives a complete assessment of current developments in deep learning-based 3D segmentation. In these approaches' benefits, draw backs, and design mechanisms are studied and addressed. This study evaluates the impact of various segmentation algorithms on competitiveness on various publicly accessible datasets, as well as the most often used pipelines, their advantages and limits, insightful findings and intriguing future research directions.

## 1 Introduction

In computer vision and graphics, segmenting 3D scenes is a dangerous and bleak issue. The purpose of 3D segmentation is to create computational algorithms that predict the fine-grained labels of objects in a 3D environment for a range of applications, such as medical image analysis, autonomous driving, industrial control, mobile, augmented and virtual reality and robotics [1,2]. The 3D segmentation can be divided into multi-sorts, instance and semantic segmentation. The goal of instance segmentation distinguishes between various instances of the same class labels in addition, semantic segmentation is to anticipate object class labels like table and chair e.g., when segmentation is able to discriminate between all individual objects, and this is referred to as instance segmentation.

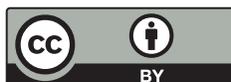

In contrast, semantic segmentation is used when segmentation can only discriminate between classes of objects. Fig. 1 is the explanation of the two segmentation techniques. Deep learning approaches have recently overtaken numerous study domains, due to their effectiveness in learning powerful features. Deep learning for 3D segmentation has also piqued the interest of the academic community throughout the last decade. However, many problems remain unresolved in 3D deep learning systems e.g., features from the RGB and depth channels are difficult to combine. The irregularity of point clouds makes exploiting local characteristics challenging, and transforming them to high-resolution voxels imposes a significant computing load [3,4]. This research provides a comprehensive summary of recent improvements in 3D segmentation using deep learning methods. It examines frequently used architectures, building components, convolution kernels and advantages and disadvantages in each scenario. Despite the fact that notable 3D segmentation surveys such as RGB-D semantic segmentation [5] and point clouds segmentation [6] have been published, these surveys do not cover all 3D data types and common application domains [2,3,5,6]. Furthermore, these surveys are mostly concerned with describing a broad review of deep learning using point clouds, rather than only 3D instance and semantic segmentation. Because of the significance of the two segmentation tasks, this research focuses solely on deep learning approaches, specifically for 3D instance and semantic segmentation. The contributions of this paper are summarized as follows:

- To our knowledge, this is the first overview study that covers deep learning approaches for 3D instance and semantic segmentation employing a variety of 3D data representations, such as RGB-D, projected pictures, voxels, point clouds, and mesh-based methods.
- Several forms of 3D instance and semantic segmentation algorithms have been carefully evaluated in terms of relative advantages and drawbacks.
- Unlike previous evaluations, we concentrate on deep learning-based algorithms for 3D instance and semantic segmentation, as well as common application domains.

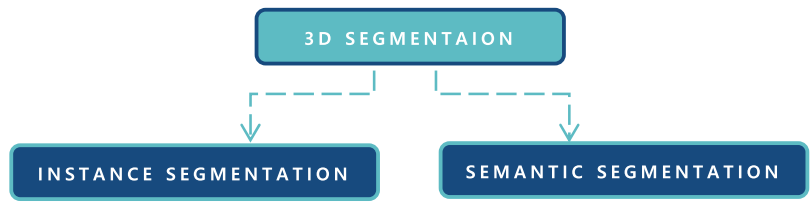

**Figure 1:** Two major divisions of 3D segmentation

Fig. 1 depicts how the rest of the article is divided into major two categories. Fig. 2 explains visual explanation about semantic and instance segmentation. Figs. 3 and 4 are explaining about their subcategories. Sections 2 and 3 covers information and underlying explanations in details, such as 3D segmentation assessment criteria. Techniques for 3D semantic segmentation are covered in Section 3, whereas methods for 3D instance segmentation are covered in Section 2. Section 4 explains the common 3D datasets and their respective 3D segmentation types, classes, sensors, points and feature representations on a variety of concise data analysis. Finally, Section 5 concludes the paper by discussing the advantages and limits and proposing future research subjects.

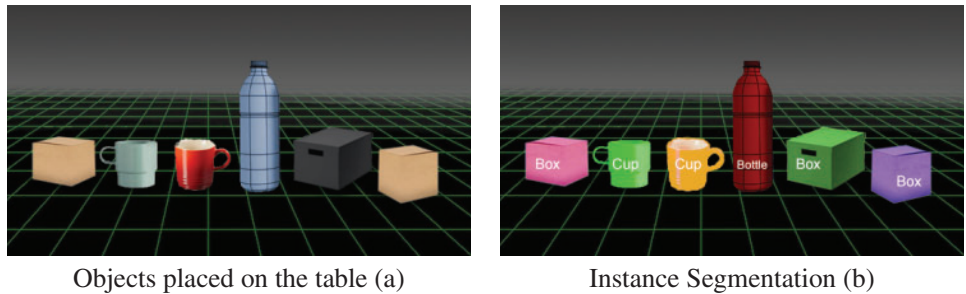

<div align="center">Objects placed on the table (a)           Instance Segmentation (b)</div>

**Figure 2:** (a) Explains about different objects on 3D plane. (b) Indicates that each instance is represented by a different color

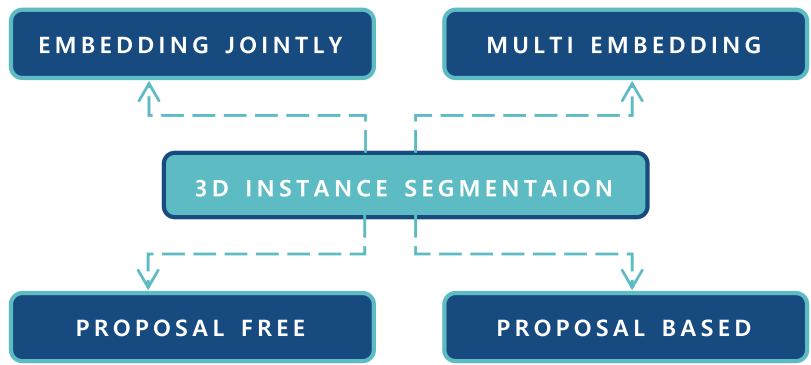

**Figure 3:** Overview of 3D instance segmentation and sub-categories

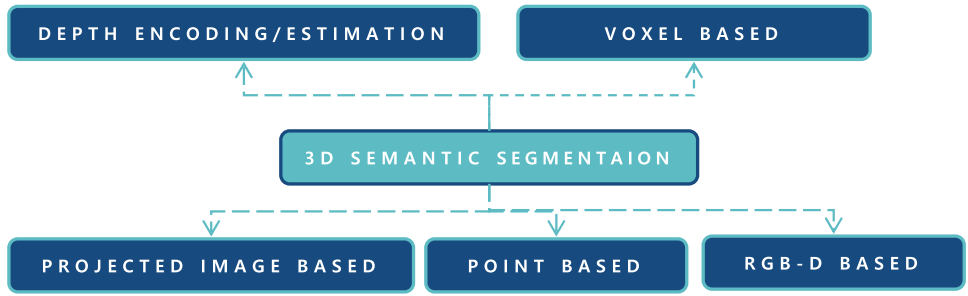

**Figure 4:** Overview of 3D semantic segmentation and its sub-categories

## 2  3D Instance Segmentation

The techniques for 3D instance segmentation distinguish between various instances of the same class. The research community is becoming more interested in 3D instance segmentation as a more informative task for scene understanding. Proposal-free and proposal-based segmentation approaches for 3D instance segmentation is essentially classified into subcategories, as explained in Fig. 1. The visual representation is explained in Figs. 2 and 5.

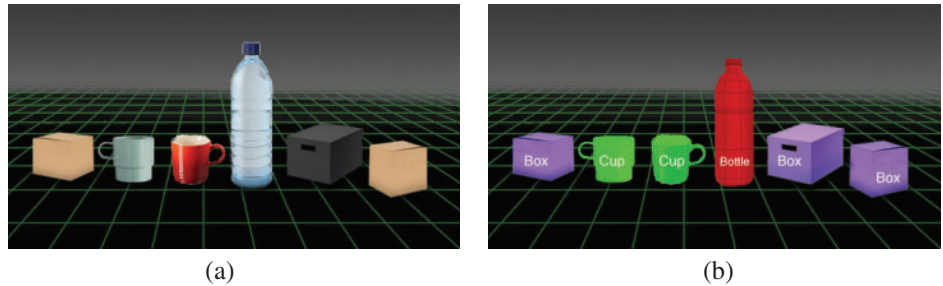

(a)                                                                (b)

**Figure 5:** Example of 3D semantic segmentation (a) represents the example of objects placed on 3D plane. (b) Represents the example of semantic segmentation with respect to object classification where each class is represented with single color example

### 2.1 Proposal Based Segmentation

The proposal-based algorithms first forecast object proposals and then refines them to build final instance masks breaking down the effort into two key issues. As a result, these strategies may be divided into detection-based and detection-free methods in terms of proposal generation.

#### 2.1.1 Detection Based Segmentation

Object suggestions are sometimes defined as a 3D bounding box regression issue in detection-based approaches. Based on the posture alignment of the 3D reconstruction, 3D-SIS combines high resolution RGB photos with voxels and simultaneously learns color and geometric characteristics using a 3D detection backbone to anticipate 3D bounding box suggestions [7]. A 3D mask backbone predicts the final instance masks in these approaches. Similarly, Pointgroup [8] proposes the Dual-set point grouping, a 3D object proposal network that reconstructs object forms from shape noise data in order to enforce geometric comprehension. The GPSN is integrated into a 3D instance segmentation network called Region based PointNet to reject, receive, and enhance proposals (R-PointNet). These networks must be trained step by step, and object proposal refinement demands an expensive suppression operation. Yang et al. [9] proposed the 3D-BoNet, a unique end-to-end network that learns a set number of 3D bounding boxes without rejection and then estimates an instance mask in each bounding box.

#### 2.1.2 Detection Free Segmentation

The detection free models including SGPN [10] is one of the detection free techniques that expects that points belonging to the same object instance should have relatively comparable characteristics. As a result, in order to forecast proposals, it learns a similarity matrix. To provide extremely credible instance ideas, the proposals are trimmed using point confidence scores. However, this basic distance similarity measure learning is ineffective in segmenting nearby items of the same class and is not informative. 3D-MPA [11] learns object proposals using sampled and aggregated point data that vote for the same object center, and then consolidates the proposal features using a graph convolutional network, which allows for higher-level interactions between proposals and refines proposal features. Jiang et al. [12] proposed a candidate assignment module and a candidate suppression module to remove superfluous candidates. In order to generate an instance grouping loss for network training, a mapping between instance labels and instance candidates is also necessary.

### 2.1.3  Proposal Free Segmentation

The methods that do not need a proposal breaking down the problem into two primary difficulties learn feature embedding for each point and then use clustering to produce definitive 3D instance labels from the embedding learning point of a view, these approaches may be loosely separated into three categories: multi-embedding learning, 2D embedding propagation and multi-task learning.

### 2.2  Multi Embedding Learning

Multi-embedding learning: approaches like MSU-Net [13] rely on good performance of the 3d semantic segmentation with submanifold sparse convolutional networks [14] to predict the similarity embedding between nearby points at different scales and semantic topology. A basic yet effective [15] is adopted to segment points into instances based on the two types of learnt embedding. The feature embedding unique to each instance and the direction embedding that orients the instance center, which provides a stronger grouping force, are both learned by multi-task metric learning (MTML) [16]. Similarly, depending on the original coordinate embedding space and the shifted coordinate embedding space Pointgroup [8] points into distinct clusters. Furthermore, the suggested ScoreNet aids in cluster selection.

### 2.3  2D Embedding Propagation Method

The 3D-BEVIS [17] approach is an example of 2D embedding, learns 2D global instance embedding with a bird's-eye view of the whole scene. The learnt embedding is then propagated onto point clouds using DGCN [18]. Another example is PanopticFusion [19], which uses the 2D instance segmentation network Mask R-CNN [20] to predict pixel-wise instance labels for RGB frames and integrates the learnt labels into 3D volumes.

### 2.4  Multi Task Jointly Learning

The impact of 3D semantic segmentation and 3D instance segmentation on each other is possible. Objects with distinct classes, for example, must be instances, but objects with the same instance label must belong to the same class. Associatively segmenting instances and semantics in point clouds (ASIS) [21] builds an encoder-decoder network called ASIS to learn semantic-aware instance embedding in order to improve the performance of the two tasks. Similarly, Joint semantic-instance segmentation of 3D point clouds with multi-task pointwise networks and multi-value conditional random fields (JSIS3D) [22] proposes an MV-CRF to simultaneously optimize object classes and instance labels by using a unified network called MT-PNet to predict the semantic labels of points and embedding the points into high dimensional feature vectors. In a similar manner, J. Du etal. [23] and Liang et el. [24] 3D-GEL uses SSCN to create semantic predictions and instance embedding at the same time, then refines the instance labels with two GCNs. OccuSeg [25] creates both an occupancy signal and a spatial embedding using a multitask learning network.

## 3  3D Semantic Segmentation

In the literature, many deep learning approaches for 3D semantic segmentation have been suggested. According on the data representation utilized, these approaches may be split into five categories: RGB-D picture based, projected images based, voxel based, point based, and other representations based. Point-based techniques are further divided into multiple layer perceptron (MLP)-based, point convolution-based, and graph convolution-based approaches, depending on the network architecture. The visual explanation presented in Fig. 5 and classification information is explained in Fig. 4 which depicts current deep learning achievements in 3D semantic segmentation.

### 3.1 RGB-D Based Segmentation

The depth map of an RGB-D picture provides geometric information about the actual environment that may be used to identify foreground items from background objects, allowing segmentation accuracy to be improved. The conventional two channel network is often used to extract features from RGB and depth pictures independently in this category. This minimal framework, however, is insufficient to extract rich and sophisticated information. To this purpose, researchers have added numerous additional modules to the aforementioned simple two channel system in order to increase performance by learning rich context and geometric information, both of which are critical for semantic segmentation. Multi-task learning, depth encoding, multi-scale networks, innovative neural network designs, data/feature/score level fusion, and post-processing are the six areas that these modules fall within (see Fig. 4).

### 3.1.1 Depth Estimation and Encoding

In computer vision, depth estimation and semantic segmentation are two fundamentally difficult tasks. Because depth variation inside an item is minor compared to depth variation across various objects, these tasks are somewhat connected. As a result, several studies combine the tasks of depth estimation with semantic segmentation. There are two basic types of multi-task leaning frameworks, cascade and parallel, based on the link between the two activities.

The depth estimate task in the cascade structure generates depth pictures for the semantic segmentation task. For example, Liu et al. [26] employed Cao et al. [27] deep convolutional neural fields (DCNF) for depth estimation. For semantic segmentation, the estimated depth pictures and RGB images are supplied into a two channel FCN. Guo et al. [28] used Ivanecky's [29] deep network for automatically producing depth pictures from single RGB photos, and then presented a two channel FCN model on the RGB and anticipated depth map image pair for pixel labeling. Because the cascade architecture does depth estimation and semantic segmentation individually, it is unable to undertake end-to-end training for two tasks at the same time. As a result, the semantic segmentation job provides no advantage to the depth estimate problem. The parallel framework, on the other hand, conducts these two activities in a network, allowing the two processes to benefit from one other. Wang et al. [30], for example, employed Joint Global CNN to offer accurate global scale and semantic guidance by using pixel-wise depth measurements and semantic labels from RGB images. They also employ Joint Region CNN to learn precise depth and semantic boundaries by extracting region-wise depth values and a semantic map from RGB.

Mousavian et al. [31] proposed a multi-scale FCN with five streams that examine depth and semantic characteristics at different sizes while sharing the underlying feature representation. To simulate the two tasks together, Liu et al. [32] suggested a collaborative de-convolutional neural network (C-DCNN). The quality of depth maps calculated from RGB photographs, on the other hand, is inferior to that obtained directly from depth sensors. In RGB-D semantic segmentation, this multi-task learning process has increasingly been abandoned. Raw depth photos provide rich geometric details that conventional 2D-CNNs are unable to explore. Another option is to convert raw depth pictures into different 2D-CNN-compatible representations. To represent the depth channel from RGB-D scenes, Höft et al. [33] employed a simplified version of the histogram of oriented gradients (HOG). From the raw depth photos, Lin et al. [34] and Pandey et al. [35] derived three additional channels: horizontal disparity, height above ground, and angle with gravity (HHA). A shortcoming of HHA, according to Liu et al. [36], is that some sceneries may not include enough horizontal and vertical planes. As a result, they suggest a new gravity direction detecting approach that employs vertical lines to learn a better representation. According to Hazirbas et al. [37], HHA representation has a high processing cost and includes less information than raw depth photos. They propose the FuseNet design, which consists of two encoder decoder branches, one for depth and the other for RGB, which directly encodes depth information with less computational effort.

### 3.1.2 Multi-Scale Networks

Small objects and detailed region segmentation benefit from the context information gained by multi-scale networks. Couprie et al. [38] used a multi-scale convolutional network to learn features directly from RGB and depth pictures using a multi-scale convolutional network. Pandey et al. [35] Introduced a multi-scale deep ConvNet for segmentation, in which coarse VGG16-FC net predictions are up sampled in a Scale-2 module and then concatenated with low-level VGG-M net predictions in a Scale-1 module to provide both high and low level features. This approach, however, is susceptible to scene clutter, resulting in output problems. Jiang et al. [39] take use of the fact that lower-resolution regions have more depth, whereas higher-resolution parts have less. They introduce context-aware receptive field (CaRF), which focuses on semantic segmentation of certain scene-resolution regions and employs depth maps to divide relevant color imagery into different scene-resolution regions. As a result, their pipeline becomes a multi-scale network.

### 3.1.3 Neural Network Based Segmentation

CNNs' capacity to process and utilize geometric information is limited because to their fixed grid calculation. As a result, additional innovative neural network designs have been developed to better utilize geometric characteristics and the correlations between RGB and depth pictures. These structures may be classified into four groups

Improved 2D Convolutional Neural Networks: Jiang et al. [39] developed a unique Dense-Sensitive Fully Convolutional Neural Network (DFCN) that includes depth information into the early layers of the network utilizing feature fusion methods, based on cascaded feature networks [34]. Following that, numerous dilated convolutional layers are used to utilize context information. Wang et al. [40] presented a depth-aware 2D-CNN by incorporating two unique layers, a depth aware convolution layer and a depth-aware pooling layer, based on the assumption that pixels with the same semantic label and comparable depth should have greater influence on one an-other. De-Convolutional Neural Networks: For the refining of segmentation maps, DCNN are a simple yet effective and efficient approach. Because of its high performance, Liu et al. [32] and Wang et al. [41] all employ the DeconvNet for RGB-D semantic segmentation. However, because the high-level prediction map combines huge context for dense prediction, DeconvNet's potential is restricted. Cheng et al. [42] developed a locality-sensitive DeconvNet (LS-DenconvNet) to enhance boundary segmentation over depth and color pictures to achieve this goal. Local visual and geometric signals from raw RGB-D data are included into each DeconvNet, allowing it to up sample coarse convolutional maps with huge con-text while recovering crisp object boundaries.

Recurrent Neural Networks: (RNNs) are capable of capturing long-distance relation-ships between pixels, however they are best suited to a single data channel (e.g., RGB). Fan et al. [43] improved single-modal RNNs to create multimodal RNNs (MM-RNNs) for RGB-D scene labeling. The MM-RNNs enable sharing of 'memory' across depth and color channels. Each channel has both its own characteristics and those of other channels, making the learnt features more discriminative for semantic segmentation. Li et al. proposed a novel Long Short-Term Memorized Context Fusion (LSTM-CF) model to gather and integrate contextual information from several channels of RGB and depth images. Li and Qi et al. [44,45] Graph Neural Networks Were the first to use GNNs for RGB-D semantic segmentation, casting 2D RGB pixels into 3D space and giving semantic information to the 3D points based on depth information. After generating a k-nearest neighbor graph from the 3D points, they employed a 3D graph neural network (3DGNN) to conduct pixel wise predictions.

### 3.1.4  Data/Feature/Score Fusion

The optimal integration of texture (RGB channels) and geometry (depth channel) story is critical for efficient semantic segmentation. Early, medium, and late fusion strategies are related to data level, feature level, and score level fusion strategies, respectively. Couprie et al. [37] concatenated the RGB and depth images into four channels for direct input to a CNN model, which is a straightforward data level fusion technique. The high connections between depth and photometric channels are not exploited by such a data level fusion. On the other hand, feature level fusion captures these relationships. For example, Li et al. [44] Developed a stored fusion layer for data driven adaptive fusing of vertical depth and RGB contexts. To retain real 2D global contexts, their technique does bidirectional propagation along the horizontal direction. Wang et al. [41] suggested a feature transformation network that connects the depth and color channels while also bridging the convolutional and de-convolutional networks in one channel. The feature transformation network can find unique features in a single channel as well as shared features across two channels, allowing the two branches to share features and increase the representation power of shared data. The sophisticated feature level fusion models mentioned are placed in a specific same layer between the RGB and depth channels, which is difficult to train and ignores other same layer feature fusion. To this purpose, Hazirbas et al. [36] perform fusion as an element-wise summing between the two channels to merge features of multiple identical layers. The simple averaging approach is often used for score level fusion. The RGB model and the depth model, on the other hand, provide distinct contributions to semantic segmentation. Liu et al. [35] developed a score level fusion layer with weighted summation that learns the weights from the two channels using a convolution layer. Cheng et al. [42] presented a gated fusion layer to learn the varied performance of RGB and depth channels in diverse scenarios for different class recognition. Both strategies outperformed the basic averaging strategy, albeit at the expense of more learnable parameters.

### 3.2  Projected Images Based Segmentation

The core idea behind projected image-based semantic segmentation is to employ 2D-CNNs to extract features from projected photos of 3D scenes/shapes and then fuse them to predict labels. Unlike a single-view picture, this pipeline not only extracts more semantic information from large-scale scenes, but it also reduces the data size of a 3D scene when compared to a point cloud. Multi-view or spherical pictures are the most common projected images.

### 3.2.1  Multiview Image Based Segmentation

To enhance classification performance, MV-CNN [46] use a unified network to aggregate data from many perspectives of a 3D shape created by a virtual camera into a single and compact shape descriptor. Researchers were motivated to apply the same concept to 3D semantic segmentation. For example, Lawin et al. [47] combine point clouds with RGB, depth, and surface normal pictures to create multi-view synthetic images. All multi-view pictures' prediction scores are combined into a single representation and back-projected onto each point. How-ever, if the density of the point cloud is poor, the snapshot may incorrectly capture the points behind the seen structure, causing the deep network to misinterpret the numerous views. Snap-Net [48,49] does this by preprocessing point clouds in order to compute point characteristics (such as normal or local noise) and generate a mesh, which is analogous to point cloud densification. They produce RGB and depth pictures from the mesh and point clouds by taking appropriate snapshots. They then use FCN to conduct pixel-wise labeling of 2D photos, and then use efficient buffering to quickly back-project these labels into 3D locations. To offer a comprehensive spatial framework for back projection, the above approaches require obtaining the whole point clouds of a 3D scene in advance. However, multi-view photographs collected directly from a real-world scene would lose a significant amount of spatial information. Some studies have attempted to combine 3D scene reconstruction with semantic segmentation, with the hope that scene reconstruction will compensate for the lack of spatial information. Boulch & Guerry et al. [48,50], for example, use global multi-view RGB and Gray stereo pictures to rebuild a 3D scene. The 2D snapshot labels are then back-projected onto the rebuilt scene. Simple back projection, on the other hand, is incapable of optimally fusing semantic and spatial geometric data. Following back projection, Pham et al. [51] suggested a revolutionary Higher-order CRF to further refine the original segmentation.

### 3.2.2 Spherical Image Based Segmentation

It's not easy to choose shots from a three-dimensional scene. To acquire an ideal depiction of the entire environment, snapshots must be made after taking into account the number of views, viewing distance, and angle of the virtual cameras. Researchers project the whole point cloud onto a sphere to avoid these difficulties. For example, Iandola et al. [52] introduced SqueezeSeg, an end-to-end pipeline based on SqueezeNet [52] that learns features from spherical pictures and refines them using CRF as a recurrent layer. PointSeg [53] enhances SqueezeNet by combining feature-wise and channel-wise attention to develop robust segmentation. SqueezeSegv2 [54] adds LiDAR mask as a channel to boost resilience to noise and improves the structure of SqueezeSeg with Context Aggregation Module (CAM). Regardless of the extent of discretization employed in CNN, RangNet++ [55] converts the semantic labels to 3D point clouds, eliminating point discarding. Despite the similarities between standard RGB and LiDAR photos, the feature distribution in LiDAR images varies depending on where you look. SqueezeSegv3 [56] uses Spatially Adaptive Convolution (SAC), a spatially-adaptive and context-aware convolution, to adopt various filters for different places.

### 3.3 Point Based Semantic Segmentation

The usage of standard 2D/3D convolutional neural networks is limited because point clouds are spread randomly in 3D space, without any canonical order and translation invariance. A group of point-based semantic segmentation networks has been suggested recently. Multiple layer perceptron (MLP)-based, point convolution based, and graph convolution-based approaches may be loosely split into three groups. There are three further sub-categories of point based semantic segmentation e.g., multiple player perceptron-based method, graph-based convolution method and point based convolution method.

Convolution operations are performed directly on the points in point convolution techniques. For example, H. Su et al. [46] uses $1 \times 1$ convolution to leverage point-wise characteristics before passing them via the local dependency module (LDM) to exploit local context features. It does not, however, specify the neighborhood for each point in order to learn about local characteristics. Li et al. [44] stacking numerous convolutional layers and a long short-term memory layer, the Long Short-Term

Memorized Context Fusion (LSTM-CF) Model merges contextual data from channels of photometric and depth data introduced to CNNs. to acquire real 2D global contexts, do bi-directional propagation of the merged vertical contexts along the horizontal way. The spatial 2D convolution neural network framework is quite similar to this method. Flex-Convolution [31] models a convolution kernel with a linear function with less parameters and adjusts inverse density importance subsampling (IDISS) to coarsen the points. For example, DA-CNN [40] uses a point probability density function (PDF) to express convolution as a Monte Carlo integration problem, with the convolution kernel represented by an MLP. Furthermore, Deep GCNs [47] uses 2D-CNN features such residual connections between layers (ResNet) to solve the vanishing gradient problem and a dilation method to allow the GCN to go deeper. Liu et al. [26] created a local graph on neighborhood points searched in multi-directions and explored local features using a local attention edge convolution, based on the fundamental architecture of PoinNet++ [57]. To capture precise and resilient local geometric details, A point-wise spatial attention module is supplied with these features. The spherical convolution kernel divides a 3D spherical area into several bins, each of which has learnable parameters for weighting the points that lie within it.

## 4 Benchmark Datasets for 3D Segmentation

The availability of public datasets has aided research on semantic segmentation's accurate border recovery. The quality of the datasets used for training is unquestionably proven by the degree of success of any deep learning-based models and applications. Only when the models are assessed against the same benchmarks are the efficiency of accurate boundary recovery strategies comparable and credible. As a result, many datasets were evaluated using the approach proposed, which are detailed in further depth. Tab. 1 shows representative 2D picture benchmark datasets for evaluating border recovery algorithms. The goal of this statistical study is to provide readers a better knowledge of the data architecture and make benchmark selection easier for future investigations.

**Table 1:** In the acquisition of these datasets for 3D semantic segmentation, many types of sensors and various 3D scanners are discussed

| Dataset | Classes | Sensors | Scenes-segmentation | Points | Feature representation |
|---|---|---|---|---|---|
| ShapeNet | 55 | - | Outdoor, indoor instance seg. | 51,300 | XYZ, Propagating human label to shapes |
| ScanNet | 21 | RGB-D | Indoor-instance seg. | 242 | XYZ, RGB, label |
| S3DIS | 13 | Structured, light | Indoor-instance seg. | 215 | XYZ, RGB, Normalized coordinates |
| PSB | 19 | Amazon's mechanical turk | Indoor-instance seg. | 380 | XYZ, segmentation, class |
| COSEG | 11 | - | Indoor-instance seg. | 1,090 | Supervised, semi-supervised |

(Continued)

**Table 1:** Continued

| Dataset | Classes | Sensors | Scenes-segmentation | Points | Feature representation |
|---|---|---|---|---|---|
| KITTI | 28 | MLS | Outdoor-semantic seg. | 1,799 | XYZ, reflectance, label, class |
| Semantic3D.net | 8 | TLS | Outdoor-semantic seg. | 4,009 | XYZ, intensity, RGB |
| Paris-Lille-3D | 50 | MLS | Outdoor-semantic seg. | 143 | XYZ, GPS time, Label, Class |
| NYUv1 & 2 | 726 | Microsoft Kinect v1 | Indoor-semantic seg. | 2,347 | 2D LabelMe-style annotation, classes |
| SUN RGB-D | 47 | RealSense, Xtion, MKv1/2 | Indoor-semantic seg. | 10,355 | 2D/3Dpolygons +3D bounding box |
| Semantic3D | 8 | Terrestrial Laser scanner | Outdoor-semantic seg. | 1,660 | XYZ, three baseline methods |
| PL3D | 50 | Velodyne HDL-32E LiDAR | Outdoor-semantic seg. | 143.1 M | Human labeling, Annotation, Class |
| Matterport3D | 90 | Matterport camera | Indoor-semantic seg. | 194.4 K | Hierarchical labeling, Annotation, Class |
| HoME & House3D | 84 | Planner5D platform | Indoor-semantic seg. | 45,622 | SSCNet +3 ways, test description |

In particular, as shown in Tab. 1, the point cloud representation is summarized, which is one of the basic methodologies for deep learning-based 3D scene interpretation. We discovered that various datasets use distinct representations, limiting their generality and appeal. If there was a consistent standard for representing point cloud features, it would undoubtedly speed up the development of more complex deep learning algorithms and their applications in the industry. Datasets are essential for training and testing deep learning based 3D segmentation algorithms. Privately gathering and annotating datasets, on the other hand, is time consuming and costly, since it necessitates subject expertise, high quality sensors, and processing equipment. As a result, relying on public datasets is an excellent approach to cut costs. Following this path has an additional benefit for the community in that it allows for a fair comparison of algorithms. Tab. 1 lists some of the most common most common datasets, organized by sensor type, data size and format, scene class, and annotation technique.

## 5 Discussion and Challenges

In this section, some of the challenges and possible solutions are discussed, followed by a comprehensive conclusion. Deep neural network (DNN) techniques for 3D instance and semantic segmentation are quickly evolving, yet the following issues remain unsolved. It is inefficient and impracticable for a system to use numerous deep learning networks to perform distinct computer vision tasks. Semantic segmentation has great consistency with various tasks, such as depth estimation, segmentation, scene comprehension, and object identification, when it comes to exploiting core features of a scene. These activities might work together to increase performance through cooperative learning.

Raw point cloud-based boundary recovery: Using numerous alternative representations, such as depth pictures, point clouds, and voxels, semantic segmentation might possibly attain improved accuracy. Single representation, on the other hand, limits segmentation accuracy due to the restrictions of scene information, such as fewer semantic, geometric, and voxel information, use of multiple representations could help to improve your performance and accuracy.

The goal of point-based semantic segmentation is to comprehensively analyze point-wise characteristics and their linkages. However, characteristics between local areas make exploitation of global context features much more difficult, and low-level features are lost.

Criterion for annotating datasets: For some applications, such as autonomous driving and mobile robotics, real-time 3D scene parsing is critical; nevertheless, most of them focus on segmentation accuracy rather than real-time speed. Few light-weight 3D semantic segmentation models, on the other hand, use pre-processing to improve segmentation speed, but they are likely to overlook a significant amount of geometric information. In the future, real-time 3D semantic segmentation approaches based on point clouds will demand greater attention.

Interpret-ability of deep learning: The neural network's output should be justified in a way that is intelligible to humans, leading to new insights into the inner workings. Interpret-able deep networks are the name given to such models. Interpret-ability isn't a one-size-fits-all concept. In reality, due to varying degrees of human comprehension, the subjectivity of an interpretation necessitates the existence of a plethora of characteristics that together define interpret-ability. Furthermore, the interpretation can be expressed either in terms of low-level network parameters or perhaps in terms of model input features. Statistical measures are commonly employed to assess an output's unpredictability. The idea of trust, on the other hand, is dependent on a human's sight into the machine's operation.

Furthermore, Due of the high computer power and resources required, present techniques are confined to extremely tiny 3D point clouds. Large-scale point clouds require data pre-processing to deal with problems like these. On the other hand, spatiotemporal characteristics can help improve the resilience of 3D video or dynamic 3D scene segmentation.

## 6 Conclusion

A detailed overview of current progresses in 3D segmentation utilizing deep learning approaches including 3D instance and semantic segmentation has been presented. Deep learning approaches for 3D segmentation have made substantial development in recent years. However, this is only the beginning, and researchers should expect major advancements in the future. In this study, various unresolved concerns were highlighted, as well as possible future topics. The papers evaluated in this study explained that effective real-time application is still a work in progress due to several limitations with point cloud data. This article on 3D point cloud segmentation, which includes a large bibliography, can give important insight into this important topic while also stimulating new study. In this research article, 3D instances and semantic segmentation methodologies are identified and explored, as well as a concise summary of their strengths and downsides. This study revealed that there is still a lot of room for development in terms of segmentation accuracy, speed, and complexity. As a result, our future work will consist involving some of these strategies and developing a new one by improving the flaws and/or combining the virtues. The issues of these complications are likely to be addressed in the near future.

**Funding Statement:** This research was supported by the BB21 plus funded by Busan Metropolitan City and Busan Institute for Talent and Lifelong Education (BIT) and a grant from Tongmyong University Innovated University Research Park (I-URP) funded by Busan Metropolitan City, Republic of Korea.

**Conflicts of Interest:** The authors declare that they have no conflicts of interest to report regarding the present study.